# An Oversampling-enhanced Multi-class Imbalanced Classification Framework for Patient Health Status Prediction Using Patient-reported Outcomes

Yang Yan, Zhong Chen, *Member, IEEE*, Cai Xu, Xinglei Shen, Jay Shiao, John Einck, Ronald C Chen, and Hao Gao

***Abstract*—Patient-reported outcomes (PROs) directly collected from cancer patients being treated with radiation therapy play a vital role in assisting clinicians in counseling patients regarding likely toxicities. Precise prediction and evaluation of symptoms or health status associated with PROs are fundamental to enhancing decision-making and planning for the required services and support as patients transition into survivorship. However, the raw PRO data collected from hospitals exhibits some intrinsic challenges such as incomplete item reports and imbalance patient toxicities. To the end, in this study, we explore various machine learning techniques to predict patient outcomes related to health status such as pain levels and sleep discomfort using PRO datasets from a cancer photon/proton therapy center. Specifically, we deploy six advanced machine learning classifiers -- Random Forest (RF), XGBoost, Gradient Boosting (GB), Support Vector Machine (SVM), Multi-Layer Perceptron with Bagging (MLP-Bagging), and Logistic Regression (LR) -- to tackle a multi-class imbalance classification problem across three prevalent cancer types: head and neck, prostate, and breast cancers. To address the class imbalance issue, we employ an oversampling strategy, adjusting the training set sample sizes through interpolations of in-class neighboring samples, thereby augmenting minority classes without deviating from the original skewed class distribution. Our experimental findings across multiple PRO datasets indicate that the RF and XGB methods achieve robust generalization performance, evidenced by weighted AUC and detailed confusion matrices, in categorizing outcomes as mild, intermediate, and severe post-radiation therapy. These results underscore the models' effectiveness and potential utility in clinical settings.**

This research was partially supported by NIH grants No. R37CA250921, R01CA261964, and a KUCC physicist-scientist recruiting grant (PI, Dr. Hao Gao).

Yang Yan and Zhong Chen are with the School of Computing, Southern Illinois University, Carbondale, IL, USA (E-mails: yang.yan@siu.edu, zhong.chen@cs.siu.edu).

Xinglei Shen, Jay Shiao, John Einck, Ronald C Chen, and Hao Gao are with the Department of Radiation Oncology, University of Kansas Medical Center, Kansas City, KS, USA (E-mails: xshen@kumc.edu, jshiao2@kumc.edu, jeinck@kumc.edu, rchen2@kumc.edu, hgao2@kumc.edu).

Cai Xu is with the Department of Bioinformatics, Border Biomedical Research Center, University of Texas at El Paso, El Paso, TX, USA (E-mail: cxu@utep.edu).

Corresponding Author: Dr. Zhong Chen, e-mail: zhong.chen@cs.siu.edu.

## I. INTRODUCTION

PATIENT reported outcomes (PROs), such as health-related quality of life, pain levels, sleep disorders, and depression feelings, have been extensively utilized at the aggregate level in both observational studies and clinical trials to gauge health and the impacts of treatment on patients [1][2]. PROs play a crucial and ongoing role in both national and international academic and clinical research, contributing to the development of thousands of reliable and valid instruments designed to measure patients' self-reported health status. Despite the numerous PRO instruments that have been developed, there remains a notable lack of focus on assessing their clinometric properties [3]. Leveraging the advanced data modeling and predictive capabilities of artificial intelligence (AI) and machine learning (ML), these technologies are emerging as a new paradigm for evaluating the reliability and validity of PRO measures [4]. This approach promises to enhance the precision and utility of PRO assessments in clinical and research settings.

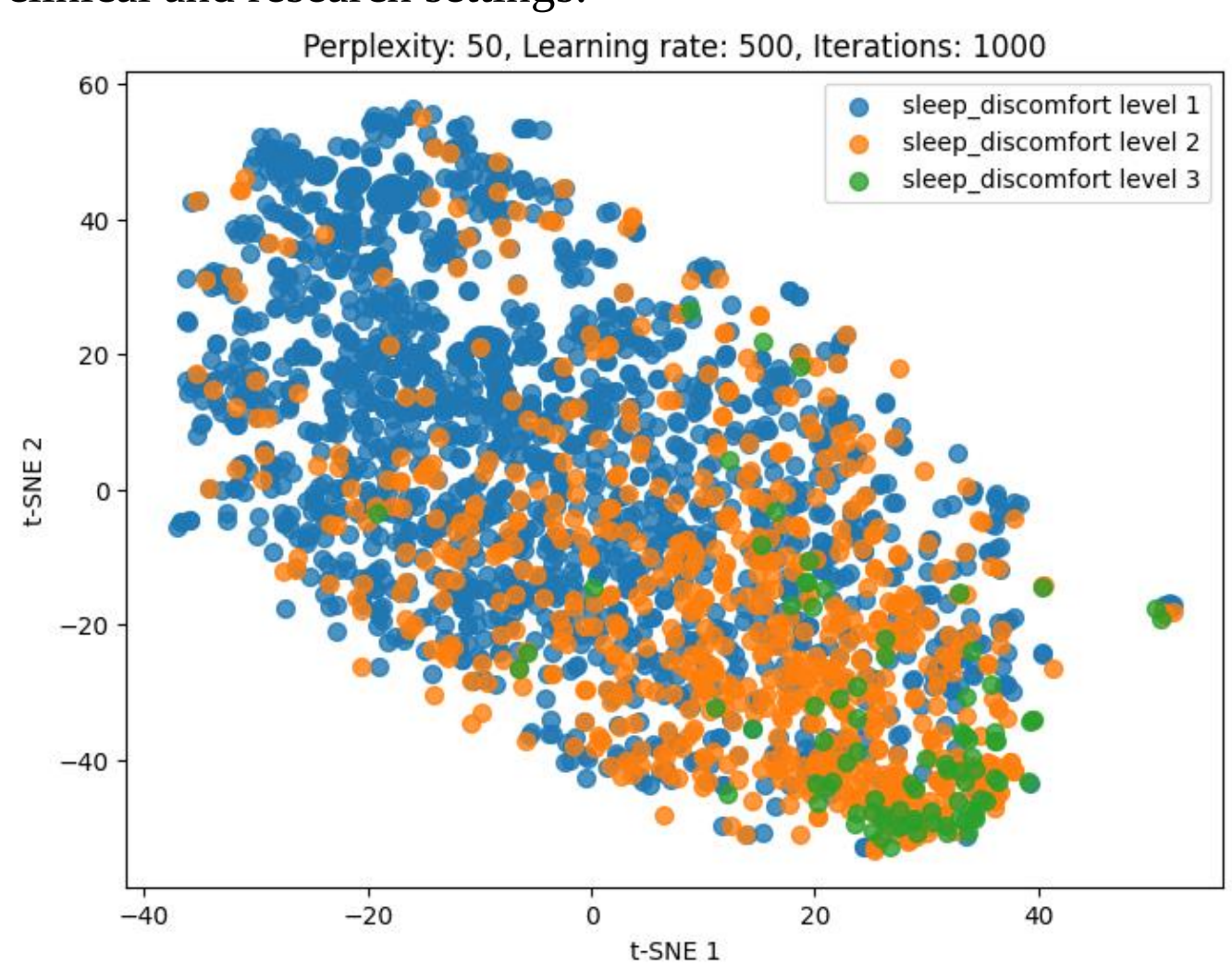

**Fig. 1.** Multi-dimensional reduction and visualization with t-SNE for breast cancer with sleeping discomfort levels 1-3. Each point represents a breast cancer patient, and each color represents the corresponding health status such as sleep discomfort level.

Although extensive psychological research has been conducted around the time of cancer diagnosis and during long-term follow-up, there remains a significant gap in understanding the impact of the end of treatments such as radiation therapy on patients' lives. Various longitudinal studies [5][6][7][8] indicate that anxiety and depression often intensify for many cancer patients following the treatment cessation. This paradoxical response can be attributed to several factors: the loss of the medical safety net, cessation of treatment as a form of active coping, reduced support from family and friends, and the fear of recurrence. Patient-reported outcomes, encompassing measures such as pain scales, discomfort, anxiety, depression, and quality of life, are crucial for understanding patient experiences before, during, and after radiation therapy. Effectively identifying patients with scaled outcomes is pivotal in clinical practice. One significant challenge is the skewed distribution of these outcomes, with far fewer patients experiencing severe pain, depression, or discomfort compared to milder cases, who may require more intense attention and healthcare. Another challenge involves patient heterogeneity, as illustrated in Fig. 1 using the multi-dimensional reduction and visualization technique t-SNE [25] for breast cancer patients. This variability spans clinical characteristics, demographics, genetic makeup, disease subtypes, treatment responses, and other health-related factors, reflecting the broader issue of intra-tumor heterogeneity common to many cancer types.

The assumption of equal misclassification costs across classes in these PRO datasets does not hold. For instance, in the context of predicting severe patient outcomes, regarding a severe outcome as the positive class and mild or intermediate outcomes as negative, missing a severe outcome (false negative) is often costlier than incorrectly identifying one (false positive) during the testing phase of PRO evaluation. Multi-class imbalance classification [9][34][35] poses several inherent challenges that can significantly impact the performance and applicability of machine learning models, especially in domains requiring high accuracy, such as PROs. The challenges include: (1) Model bias toward majority classes [36][37]. Models trained on imbalanced datasets tend to favor the majority classes. This occurs because many standard learning algorithms are designed to maximize overall accuracy, often at the expense of the minority classes; (2) Overfitting to majority classes [38][39][40]. There is an increased risk of overfitting to the majority classes, as the model is exposed to more examples from these groups, potentially leading to the learning of noise rather than generalizable patterns. This undermines the model’s ability to perform well on new, particularly minority class data; and (3) Decreased sensitivity for minority classes [41][42][43]. The sensitivity (true positive rate) for minority classes is typically much lower due to fewer examples to learn from. This leads to poor recognition of these critical classes, which may have severe consequences in contexts like disease detection as identifying rare diseases accurately is vital.

Thanks to rapid advancements in AI and ML, numerous sophisticated algorithms have been developed to address the challenges of multi-class imbalanced classification [9][34][35]. Generally, strategies to mitigate class imbalance can be categorized into three groups: data-level approaches, algorithm-level approaches, and ensemble learning approaches. Each category offers distinct methods to tackle the aforementioned challenges, enhancing the efficacy and reliability of models in critical applications.

Data-level approaches address the class imbalance by adjusting the data distribution through resampling methods. These techniques include increasing the size of the minority class or decreasing the majority class observations. Various resampling strategies, such as random over-sampling, random under-sampling, and the Synthetic Minority Over-sampling Technique (SMOTE) [10], have been developed, offering distinct advantages and drawbacks. For instance, effective sampling can balance the dataset to enhance machine learning model training. However, biased sampling instances may lead to degraded performance.

Algorithm-level approaches adapt the ML algorithms to better handle imbalanced data [22]. These methods identify and modify elements within the training procedure that contribute to bias toward the majority class, rendering these mechanisms insensitive to data with skewed distributions. Considering internal modifications, their effectiveness depends on the specific problem and classifier involved. A common method involves directly altering the learning procedure of an algorithm. Cost-sensitive solutions [29][30][31][32][43], another set of algorithm-level strategies, assign higher costs to misclassifying minority class instances, thereby penalizing errors more heavily on less represented classes. This approach aims to minimize classification cost and create a more equitable decision boundary.

Ensemble classifiers [26][27][28], also known as multiple classifier systems, enhance learning performance by combining several base classifiers. Each classifier's output is aggregated to form a composite decision for new samples. In boosting, for example, a sequential aggregate of base classifiers is constructed using weighted versions of the training data, with a focus on samples misclassified in previous stages. The weights of the samples are adjusted based on each classifier's performance, optimizing the ensemble's overall accuracy.

Additionally, adequately assessing PROs may require complete and sufficient collection of data beyond standard clinical measures. However, patients may have concerns such as security and privacy by providing complete PRO items. It is hence another big challenge to learn from incomplete PRO datasets. To ensure accurate and high-quality assessment of PRO-related symptoms, this study explores a range of machine learning classifiers tailored to address the challenges of multi-class imbalanced classification. Specifically, we first utilized the iterative imputation method, which benefits from learning conditional distributions directly. We then use label encoding and standard scaling to normalize the data. We finally applied an oversampling strategy to adjust the sample sizes of representative classes within the training set, ensuring equitable classification across all classes during training. The core principle is that the interpolations for the imbalanced training sets remain true to the original skewed class distribution even after oversampling. Extensive experiments were conducted across prostate, head and neck, and breast cancer cases utilizing six widely used ML classifiers. Our findings indicate that RF and XGBoost maintain superior classification performance and precise feature interpretations compared to other baseline classifiers, demonstrating strong potential for real-world clinical application.

The main contributions of our research to the field of PRO can be summarized as follows.

- We provide a comprehensive evaluation of a set of classifiers -- including Support Vector Machine (SVM), Logistic Regression (LR), Gradient Boosting (GB), XGBoost (XGB), Random Forest (RF), and Bagging with Multi-Layer Perceptron (Bagging-MLP) -- alongside an oversampling strategy to tackle multi-class imbalance in PRO datasets.
- We detail the feature interpretations related to PRO, derived from questionnaires administered before, during, and after radiation therapy.
- We report on the classification performance and time efficiency of top ML classifiers, offering insights that guide their clinical use.

The rest of the paper is organized as follows. The related work is discussed in Section II. The utilized approaches are elaborated in Sections III. The experimental results are reported in Section IV. The conclusion is summarized in Section V.

## II. Related Work

In this section, we review the existing exploratory studies on PRO, and state-of-the-art ML for PRO prediction.

### *A. Existing exploratory studies on PRO with radiation therapy*

PROs [46] have gained significant attention in radiation therapy research due to their ability to capture patients' subjective experiences with treatment, helping clinicians better evaluate chemotherapy, immunotherapy, or radiotherapy toxicity. Grover et al. [47] investigated a study to understand patient-reported toxicities resulting from treatment of prostate cancer using various different modalities that have similar oncological endpoints. They found that men with prostate cancer experience significant urinary and sexual sequelae from treatment regardless of the modality used. Patients treated with surgery reported more urinary and sexual side effects than those treated with radiation. The goal of the study in [48] is to design a confirmatory factor analysis for finding clinically relevant symptom groups from PROs and to test associations between symptom groups and radiation dose.

Barocas et al. [49] investigated functional outcomes and adverse effects associated with radical prostatectomy, external beam radiation therapy (EBRT), and active surveillance. In the prospective, population-based cohort study involving 2550 men, radical prostatectomy was associated with significant declines in sexual function compared with external beam radiation therapy and active surveillance at 3 years. Arbab et al. [50] conducted an exploratory study that aims to identify subgroups of patients with cancer receiving radiation therapy who have different mental health and well-being trajectories, and examine which socio-demographic, physical symptoms, and clinical variables are associated with such trajectories. They found that respiratory cancer was the most common diagnosis with moderate to severe comorbidity burden. Four latent classes with distinct anxiety, depression, and well-being trajectories were identified.

### *B. State-of-the-art machine learning methods for PRO*

ML has increasingly been integrated into the analysis of PROs to enhance predictive accuracy, personalize treatment, and improve overall patient care in radiation therapy and other medical fields. For example, Staartjes et al. [11] trained deep neural network-based and logistic regression-based prediction models for patient-reported outcome measures. The primary endpoint was achievement of the minimum clinically important difference (MCID) in numeric rating scales and Oswestry Disability Index, defined as a 30% or greater improvement from baseline. A total of 422 patients were included. After 1 year, 337 (80%), 219 (52%), and 337 (80%) patients reported a clinically relevant improvement in leg pain, back pain, and functional disability, respectively. The deep learning models predicted MCID with high area-under-the-curve of 0.87, 0.90, and 0.84, as well as accuracy of 0.85, 0.87, and 0.75.

Tschuggnall et al. [51] apply modern machine learning techniques to a real-life dataset consisting of data from more than a thousand rehab patients (N=1,047) and build models that are able to predict the rehab success for a patient upon treatment start. By utilizing clinical and patient-reported outcome measures (PROMs) from questionnaires, they compute patient-related clinical measurements (CROMs) for different targets like the range of motion of a knee, and subsequently use those indicators to learn prediction models. Pfob et al. [52] trained, tested, and validated three machine learning algorithms (LR with elastic net penalty, XGBoost, and neural network) to predict clinically important differences in satisfaction with breasts at 2-year follow-up using the validated BREAST-Q. Sidey et al. [53] collected PRO data from a single academic cancer institution in the US. Women with recurrent ovarian cancer completed biopsychosocial PROMs every 90 days. They randomly partitioned the dataset into training and testing samples with a 2:1 ratio. They used synthetic minority oversampling to reduce class imbalance in the training dataset. They fitted training data to six machine learning algorithms and combined their classifications on the testing dataset into a voting ensemble.

Yang et al. [54] used machine learning algorithms and statistical models to explore the connection between radiation treatment and post-treatment gastro-urinary function. Since only a limited number of patient datasets are currently available, they used image flipping and curvature-based interpolation methods to generate more data to leverage transfer learning. Using interpolated and augmented data, they trained a convolutional autoencoder network to obtain near-optimal starting points for the weights. A convolutional neural network then analyzed the relationship between patient-reported quality-of-life and radiation doses to the bladder and rectum. They also used analysis of variance and logistic regression to explore organ sensitivity to radiation and to develop dosage thresholds for each organ region. Their findings show no statistically significant association between the bladder and quality-of-life scores. However, they found a statistically significant association between the radiation applied to posterior and anterior rectal regions and changes in quality of life.

## III. Methodology

### A. PRO Data Collection and Pre-processing

In this study, we concentrated on analyzing the three most prevalent cancer types within our dataset: prostate, head and neck, and breast cancer. We began by identifying the largest cancer categories through a count of unique cases, ultimately selecting the top three for in-depth analysis. Our preprocessing steps included handling missing data by removing rows and columns with null or NaN values. To address the incomplete and class-imbalance issues, we first utilized the iterative imputation method, which benefits from learning conditional distributions directly. We then use label encoding and standard scaling to normalize the data. Lastly, we use a random oversampling strategy, to increase the representations of particularly for the minority classes.

We provide a comprehensive evaluation of a set of classifiers--including Logistic Regression (LR), Support Vector Machine (SVM), Gradient Boosting (GB), XGBoost (XGB), Random Forest (RF), and Bagging with Multi-Layer Perceptron (Bagging-MLP)--alongside an oversampling strategy to tackle multi-class imbalance in PRO datasets. LR uses a sigmoid function to predict the probabilities of each class. SVM uses kernel tricks to enhance prediction ability. XGBoost is an optimized distributed gradient boosting method based on decision trees designed to be highly efficient and flexible. MLP uses the binary cross entropy loss to update network parameters.

To optimize model performance, we employ grid search with cross-validation to fine-tune hyper-parameters for the six ML algorithms. The grid search will explore a range of hyper-parameters in each ML model, which is to search of all the parameters in steps to obtain the best ones. Through this methodology, our goal is to identify the model that delivers the most accurate classification results.

### B. Logistic Regression (LR)

Logistic Regression (LR) [12] is a linear statistical model that employs a logistic function to model a binary dependent variable based on independent variables. One of LR's key strengths is its high interpretability as a linear model. Additionally, it provides a probability estimate for an input vector's membership in a particular class, which is instrumental in prioritizing critical outcomes, such as severe pain, diminished quality of life, and deep feelings of depression following radiation therapy. The benefit of the LR algorithm is that variables do not require normal distribution. Independent variables in the LR could be designated as 0 and 1, denoting the outcome absence and presence. Model output varies between 0 and 1 and represents outcome susceptibility. The LR is based on the logistic function $P$, determined as $P = \frac{1}{1+e^{-X}}$, where $P$ denotes the probability related to a certain observation, and $X$ could be defined as $X = \beta_0 + \beta_1 x_1 + \beta_2 x_2 + \cdots + \beta_n x_n$ , where $\beta_0$ denotes the intercept of the algorithm, $\beta_i$ is the coefficient representing the independent variables contribution $x_i$ , and $n$ denotes the number of conditioning factors. Given these attributes, we will include LR as one of the baselines in our comparison of multi-class classification models.

### C. Multilayer Perceptron-Bagging (MLP-Bagging)

The Multilayer Perceptron (MLP) is a type of feed-forward neural network [13] based on Perceptrons that are linear combinations of input and weights. In an MLP, neurons are arranged in layers and fully connected, meaning each neuron in one layer connects to every neuron in the subsequent layer. This architecture allows the network to map input data through multiple layers of neurons, effectively utilizing hidden layers to process and transform the data. The predominant learning technique used in MLPs is back-propagation, where the network minimizes output errors by adjusting the weights of the neuron connections iteratively across each layer during the learning process.

Bagging-MLP, a variant of the standard MLP, is an ensemble learning method that enhances the stability and accuracy of machine learning algorithms [14]. The core strategy of bagging (Bootstrap Aggregating) involves creating multiple versions of a training dataset through random sampling with replacement, training a separate model on each set, and then combining their outputs. This approach reduces the variance of model prediction and helps avoid overfitting as well, making the Bagging-MLP particularly effective for handling complex datasets where a single model might struggle to perform consistently.

### D. Support Vector Machine (SVM)

Support Vector Machines (SVM) [15] are renowned for their ability to optimize expected learning outcomes. Originally introduced as a kernel-based machine learning model, SVMs are employed extensively in classification and regression tasks. Their exceptional generalization capabilities, combined with optimal solutions and discriminative power, have garnered significant attention from data mining, pattern recognition, and machine learning communities in recent years. SVMs operate by constructing decision functions directly from the training data, aiming to maximize the separation (margin) between decision boundaries in a high-dimensional feature space. This approach not only minimizes the classification errors in the training data but also enhances the model's generalization ability. SVMs are particularly noted for their superior performance in comparison to other supervised learning methods, especially in scenarios involving a small number of input data.

A notable feature of SVMs is their use of support vectors--selected subsets of the training data that define the decision boundary. These support vectors represent only a small portion of the entire dataset but are critical in defining the classification task. This efficient use of data helps SVMs to achieve robust performance in both data classification and regression analysis, making them one of the most favored techniques in machine learning.

### *E. Gradient Boosting (GB)*

The Gradient Boosting (GB) method, as outlined in recent studies [16], employs multiple additive trees (MATs) and serves as a robust machine learning and data mining technique for tackling both regression and classification problems. This method is an enhancement over the simple decision tree (DT) model, utilizing stochastic gradient boosting to incrementally improve prediction accuracy.

GB operates by constructing an ensemble of decision trees, where each tree incrementally corrects the errors made by its predecessors. This approach not only retains all the benefits of decision tree models, such as ease of interpretation and handling of heterogeneous features but also enhances their performance by increasing robustness and reducing susceptibility to overfitting. Key advantages of the GB model include its ability to process large datasets without the need for preprocessing, its resistance to outliers, and its capability to handle missing data effectively. Furthermore, its robustness allows it to manage complex data structures reliably.

Functionally, a GB model approximates the true relationship within the data through a series expansion. It begins with a simple decision tree to establish a baseline level of prediction error. Subsequent trees are then trained on the residuals or errors of the preceding tree, progressively refining the model. This iterative process continues until the improvements become negligible, thereby optimizing the model's performance and minimizing errors. This methodical refinement makes GB a powerful tool in the arsenal of predictive analytics.

### *F. XGBoost*

XGBoost [16] utilizes decision trees as the base classifiers for constructing an ensemble. However, XGBoost is distinguished by its use of gradient boosting, where the assembly of simple decision trees into a robust ensemble is directed by the optimization of a differentiable loss function. In this boosting framework, the classifier outputs class labels by applying a logistic transformation to the linear combination of individual decision tree outputs, effectively predicting probabilities.

XGBoost enhances traditional gradient-boosting methods through several key improvements, enabling a highly scalable tree-boosting system. Noteworthy enhancements include the introduction of regularization techniques to prevent overfitting, the ability to handle sparse and weighted data effectively, and the adoption of a block structure to facilitate parallel processing. These features collectively contribute to the robustness and efficiency of XGBoost, making it a preferred choice for tackling complex machine learning challenges.

### *G. Random Forest (RF)*

A Random Forest (RF) classifier [17] is an ensemble approach that constructs multiple decision trees, each based on a randomly selected subset of training samples and features. This method has gained widespread popularity in data mining and machine learning communities due to its high classification accuracy. In RF, each tree is developed through a process of repetitive partitioning starting from the root node, with the same node splitting procedure applied until certain stopping rules are met. The strength of RF lies in the aggregation of these many weaker learners (decision trees), which collectively enhance prediction accuracy. The performance of RF is particularly robust when the correlations between individual trees in the forest are low.

In binary decision trees, the node-splitting process involves selecting a variable and establishing a rule for the split. This process is guided by the goal of minimizing node impurity, typically measured using the Gini index for categorical response variables or variance for quantitative responses. The growth of each decision tree ceases when the nodes are pure (all samples within a node belong to the same class or share the same response value), or other predefined stopping criteria are achieved. The terminal nodes, or leaves, are then used to predict new observations.

To predict with RF, an observation is passed through every decision tree in the forest. The final prediction is derived by either majority voting or averaging the results across all trees. Since RF utilizes bootstrap sampling to develop each tree, some observations are not included in each tree’s construction. Furthermore, RF allows for the evaluation of variable importance, providing a measure of each predictor's relevance to the prediction process.

### *H. Oversampling*

Oversampling approaches [18][19][20][21] are designed to augment the minority class in datasets by creating copies of existing samples or introducing additional samples. These methods expand the training set size, which can lead to increased computational time. The simplest form of increasing the minority class size is random oversampling, which duplicates existing minority class samples randomly but often leads to overfitting.

To mitigate these issues, the Synthetic Minority Over-sampling Technique (SMOTE) [10] was developed. SMOTE generates synthetic instances by creating linear combinations of pairs of similar samples from the minority class, thereby providing new information to enhance the learning algorithm's performance on minority class predictions.

The core innovation of SMOTE is the creation of synthetic examples through interpolation among several minority class instances located within a specific neighborhood. This approach focuses on the "feature space"--the attributes and their interactions--rather than merely duplicating data in the "data space". This method necessitates a deeper theoretical exploration of the relationships between original and synthetic instances, considering factors such as data dimensionality. Additionally, it's important to study the variance, correlation, and relationship between the distributions of training and test examples in both the data and feature spaces. These considerations are crucial for understanding and improving the effectiveness of SMOTE and similar oversampling techniques [45].

Hence, we utilized SMOTE in this study to create synthetic positive examples for minority classes. After resampling, the above conventional (cost-insensitive) ML algorithms can then be applied on the rebalanced training datasets. One advantage of this strategy is that it is applicable to any existing classification algorithms. Such oversampling techniques can also be incorporated within ensemble learning algorithms and have received much attention recently due to their convincing performance on imbalanced datasets.

TABLE I

SUMMARY OF THE PRO DATASETS UTILIZED IN THE EXPERIMENTS.

| Dataset | #Attributes | #Training Instances | #Training Instances (Oversampling) | #Testing Instances | #Classes |
|---|---|---|---|---|---|
| Head&Neck (pain) | 25 | 2,525 (939:657:554:304:71) | 4,695 (939:939:939:939:939) | 632 (237:158:138:80:19) | 5 |
| Prostate (bowel pain) | 22 | 3,139 (1,477:767:491:339:65) | 7355 (1,477:1,477:1,477:1,477:1,477) | 784 (362:208:123:79:12) | 5 |
| Breast (sleeping) | 24 | 2,054 (1,430:555:69) | 4,290 (1,430:1,430:1,430) | 514 (361:131:22) | 3 |

## IV. EXPERIMENTS

We conducted a series of experiments to verify the effectiveness of the ML approaches for the PRO data analysis.

### *A. Datasets*

Table I presents the number of cases for each cancer type and the class distribution within the datasets. The head and neck dataset consists of 25 attributes, and 2,525 training patients distributed across five classes. The notation (939:657:554:304:71) indicates the number of cases per class, with 939 patients in class 1, 657 patients in class 2, 554 patients in class 3, 304 patients in class 4, and 71 patients in class 5, respectively. After applying random oversampling to balance the class distribution, the training set increased to 4,695 patients, with equal representation across classes.

Similarly, the prostate dataset contains 22 attributes, and 3,139 training patients spread across five classes, ranging from 1,477 patients in class 0 to 65 patients in class 4 (1,477:767:491:339:65). After oversampling, the training set expanded to 7,355 patients, ensuring equal representation for all classes. The test set consists of 784 patients, distributed as (362:208:123:79:12).

The breast dataset includes 24 attributes, and 2,054 training patients distributed across three classes, with class 1 having 1,430 patients and class 3 having 69 patients (1430:555:69). Following oversampling, each class contains 1,430 patients, totaling 4,290 training patients, while the test set has 514 patients distributed as (361:131:22).

This oversampling process was essential for balancing the datasets, reducing bias toward majority classes, and improving model performance across all cancer types.

### *B. Evaluation Metrics*

We utilized precision, recall, F1-score, AUC score, ROC, confusion matrix, and average running time [23][24] in multi-class imbalance classification as the evaluation metrics to measure the performance of all completing algorithms.

TABLE II

THE PARAMETERS OF TRADITIONAL MACHINE LEARNING ALGORITHMS AND THEIR PARAMETER VALUE RANGES.

| Method | Group | Number of subjects |
|---|---|---|
| RF | N estimators | [10, 50, 100, 200] |
| | Max features | [sqrt, log2] |
| | Max depth | [None, 10, 20, 30, 40, 50] |
| | Min samples split | [2, 5, 10] |
| | Max samples leaf | [1, 2, 4] |
| XGB | N estimators | [50, 100, 200] |
| | Learning rate | [0.01, 0.1, 0.2] |
| | Max depth | [3, 4, 5] |
| | Min child weight | [1, 2, 3] |
| | Gamma | [0, 0.1, 0.2] |
| | Subsample | [0.8, 0.9, 1.0] |
| | Colsample bytree | [0.8, 0.9, 1.0] |
| GB | N estimators | [50, 100, 200] |
| | Learning rate | [0.01, 0.1, 0.2] |
| | Max depth | [3, 4, 5] |
| | Min samples split | [2, 4] |
| | Max samples leaf | [1, 2] |
| | Max features | [sqrt, log2, None] |
| SVM | Kernel | rbf |
| | C | 10 |
| | Decision function shape | ovr |
| | Gamma | 0.01 |
| MLP-Bagging | Max iteration | 1000 |
| | Initial learning rate | 0.001 |
| | Solver | adam |
| | Estimator | base_mlp |
| | N estimators | 10 |
| LR | C | [0.1, 1, 10, 100, 1000] |
| | Solver | [newton-cg, lbfgs, sag, saga] |
| | Max iterations | [100, 200, 300, 500, 1000] |

(1) **Precision** measures the accuracy of positive predictions. It is the ratio of true positive (TP) predictions to the sum of true positive and false positive (FP) predictions, $precision = \frac{TP}{TP+FP}$. High precision indicates that the model has a low false positive rate.

(2) **Recall** measures the model's ability to identify all relevant instances. It is the ratio of true positive (TP) predictions to the sum of true positives and false negatives (FN), $recall = \frac{TP}{TP+FN}$. High recall indicates that the model captures most of the positive instances.

(3) **F1-score** is the harmonic mean of precision and recall, providing a balance between the two metrics, $F1 - score = 2 = \frac{precision \cdot recall}{precision+recall}$. It is particularly useful when dealing with imbalanced datasets, as it considers both false positives and false negatives.

(4) **AUC (Area Under the Curve)** represents the area under the ROC curve, providing a measure of the model's ability to distinguish between classes. An AUC of 1 indicates perfect classification, while an AUC of 0.5 indicates no discrimination capability.

(5) **ROC (Receiver Operating Characteristic)** Curve is a graphical representation of the true positive rate (sensitivity) against the false positive rate at various threshold settings. The ROC curve helps to visualize the trade-off between sensitivity and specificity.

(6) **Confusion Matrix** is used to describe the performance of a classification model. It shows the true positives, true negatives, false positives, and false negatives, allowing for a detailed analysis of the model's performance.

(7) **Running Time.** The average running time of different

classification methods is reported in the experiments.

### *C. Implementation Details*

For handling missing values in numerical columns, we employ the IterativeImputer method from the sklearn.impute library. Additionally, we use math.ceil() to round the imputed values to the nearest integer. The dataset is split into training and testing sets using five-fold cross-validation. For oversampling the training data, we apply the RandomOverSampler from the imblearn library. To optimize model performance, we employ GridSearchCV from sklearn.model_selection to fine-tune hyper-parameters for various ML algorithms, including SVM, LR, GB, XGB, and RF. The grid search will explore a range of hyper-parameters in each ML model, which is an exhaustive search of all the parameters in steps to obtain the best ones. The parameter settings of the ML models we selected are shown in Table II.

### *D. Overall Comparison*

For three cancer datasets featuring diverse PRO outcomes, the comparative performance of all competing algorithms in terms of precision, recall, F1-score, and weighted AUC score is summarized in Table III. Several observations regarding performance can be made. Firstly, Random Forest (RF) consistently outperforms the other five baseline models in most of the datasets with the highest F1-score and AUC score. This underscores RF's attractiveness as a method due to the simplicity of its computations, its efficacy in multi-class problems, and its high performance in terms of F1-score. Additionally, RF's robustness is enhanced by its ensemble nature.

TABLE III

PRECISION, RECALL, F1-SCORE, AND WEIGHTED AUC-SCORE COMPARISON OF SIX MACHINE LEARNING CLASSIFIERS ON THREE CANCER PRO DATASETS. THE BEST RESULTS ON EACH DATASET ARE HIGHLIGHTED IN BOLD.

| Method | Metric | H&N (pain) | H&N (QoL) | Pro (bowel pain) | Pro (depression) | Pro (sexual) | Breast (muscle) | Breast (sleeping) |
|---|---|---|---|---|---|---|---|---|
| RF | Precision | 0.533 | 0.498 | 0.702 | **0.454** | 0.661 | **0.787** | 0.750 |
| | Recall | **0.553** | 0.494 | 0.696 | **0.460** | **0.672** | **0.785** | **0.740** |
| | F1-score | **0.541** | **0.494** | 0.697 | **0.455** | **0.662** | **0.784** | 0.744 |
| | AUC | **0.815** | 0.776 | **0.902** | **0.761** | **0.883** | **0.845** | **0.848** |
| XGB | Precision | 0.527 | 0.497 | 0.691 | 0.419 | 0.660 | 0.756 | **0.755** |
| | Recall | 0.542 | 0.488 | 0.691 | 0.417 | 0.665 | 0.759 | 0.739 |
| | F1-score | 0.536 | 0.488 | 0.692 | 0.416 | 0.66 | 0.76 | **0.741** |
| | AUC | 0.794 | **0.785** | 0.898 | 0.734 | 0.878 | 0.829 | 0.827 |
| GB | Precision | 0.524 | 0.491 | 0.688 | 0.406 | 0.639 | 0.741 | 0.731 |
| | Recall | 0.535 | **0.496** | 0.693 | 0.411 | 0.646 | 0.748 | 0.737 |
| | F1-score | 0.526 | 0.492 | 0.688 | 0.407 | 0.641 | 0.746 | 0.735 |
| | AUC | 0.794 | 0.761 | 0.891 | 0.719 | 0.867 | 0.806 | 0.815 |
| SVM | Precision | 0.526 | **0.518** | **0.722** | 0.420 | **0.671** | 0.744 | 0.764 |
| | Recall | 0.521 | 0.485 | **0.705** | 0.422 | 0.637 | 0.746 | 0.730 |
| | F1-score | 0.519 | 0.492 | **0.711** | 0.419 | 0.647 | 0.756 | 0.740 |
| | AUC | 0.769 | 0.783 | 0.865 | 0.726 | 0.850 | 0.802 | 0.819 |
| MLP-Bagging | Precision | 0.522 | 0.483 | 0.696 | 0.429 | 0.649 | 0.750 | 0.730 |
| | Recall | 0.525 | 0.472 | 0.690 | 0.427 | 0.649 | 0.741 | 0.727 |
| | F1-score | 0.522 | 0.476 | 0.694 | 0.427 | 0.649 | 0.750 | 0.726 |
| | AUC | 0.778 | 0.758 | 0.886 | 0.720 | 0.863 | 0.813 | 0.805 |
| LR | Precision | **0.544** | 0.469 | 0.698 | 0.339 | 0.649 | 0.757 | 0.769 |
| | Recall | 0.524 | 0.418 | 0.681 | 0.343 | 0.631 | 0.722 | 0.704 |
| | F1-score | 0.528 | 0.425 | 0.686 | 0.327 | 0.630 | 0.739 | 0.723 |
| | AUC | 0.801 | 0.719 | 0.757 | 0.683 | 0.780 | 0.719 | 0.837 |

Secondly, RF also exhibits relatively higher precision and recall across most datasets, with specific exceptions noted in the head and neck dataset for quality-of-life outcomes, and the prostate dataset for bowel pain outcomes. This suggests a lower rate of false positives and false negatives with RF. Conversely, the highest precision rates are recorded by Support Vector Machine (SVM) in the head and neck (Quality of Life) and prostate (Bowel Pain and Sexual Quality) datasets. This indicates that both RF and SVM potentially reduce the number of false positives in such imbalanced multi-class classification scenarios, offering significant promise for managing severe outcomes in cases with fewer patients.

TABLE IV

TRAINING TIME (SECONDS) COMPARISON OF SIX MACHINE LEARNING CLASSIFIERS ON THREE CANCER PRO DATASETS.

| Method | RF | XGB | GB | SVM | MLP-Bagging | LR |
|---|---|---|---|---|---|---|
| Head&Neck (pain) | 353.18 | 3815.68 | 1972.74 | 0.35 | 825.15 | 11.49 |
| Prostate (bowel pain) | 594.53 | 2167.71 | 3513.67 | 0.54 | 1039.73 | 26.71 |
| Breast (sleeping) | 301.17 | 909.25 | 1071.57 | 0.19 | 691.33 | 9.87 |

For the running time comparison in Table IV, we found that the SVM algorithm was the most efficient algorithm, achieving the lowest computational cost among the tested models across three distinct cancer types. This efficiency is followed by LR, RF, MLP-Bagging, XGB, and B, in order. The relatively higher computational times for RF, MLP-Bagging, XGB, and GB can primarily be attributed to their use of grid search strategies for hyper-parameter optimization, which involves exhaustively searching through every possible combination of hyper-parameter values to find the best settings. This thorough search contributes to their increased running times compared to SVM and LR.

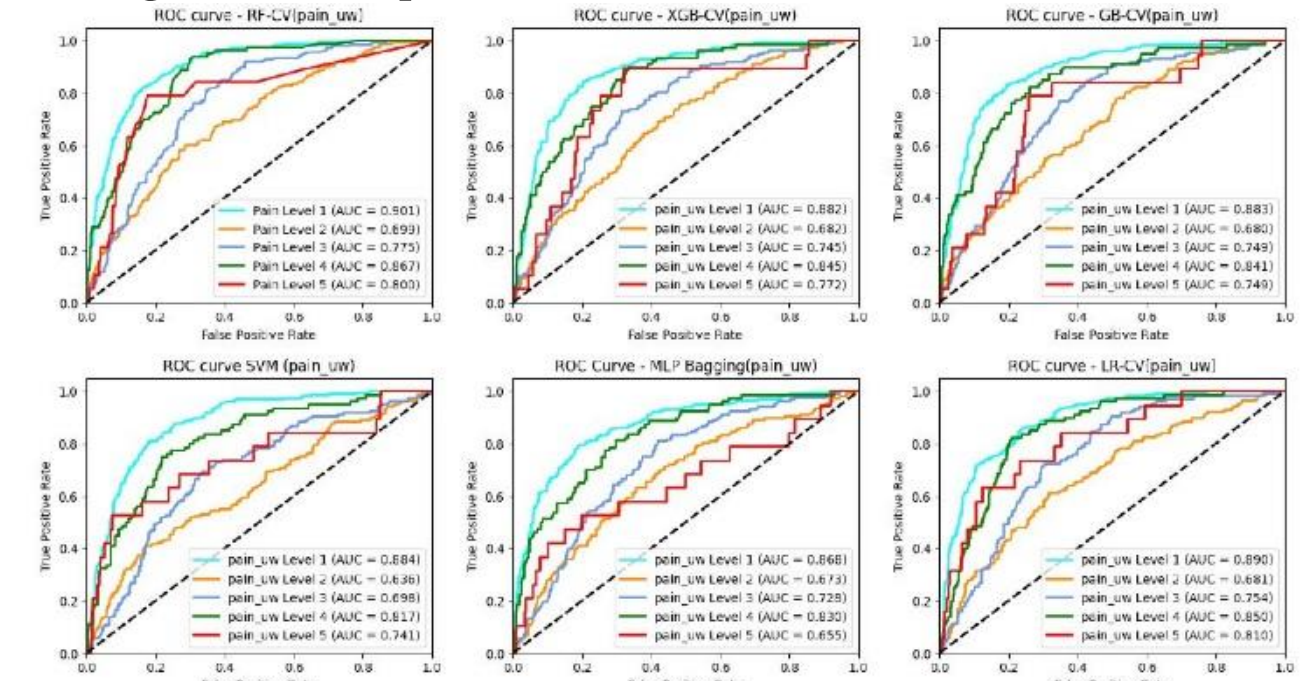

**Fig. 2.** AUC curves of six different machine learning classifiers on the head and neck cancer PRO (pain level 1-5).

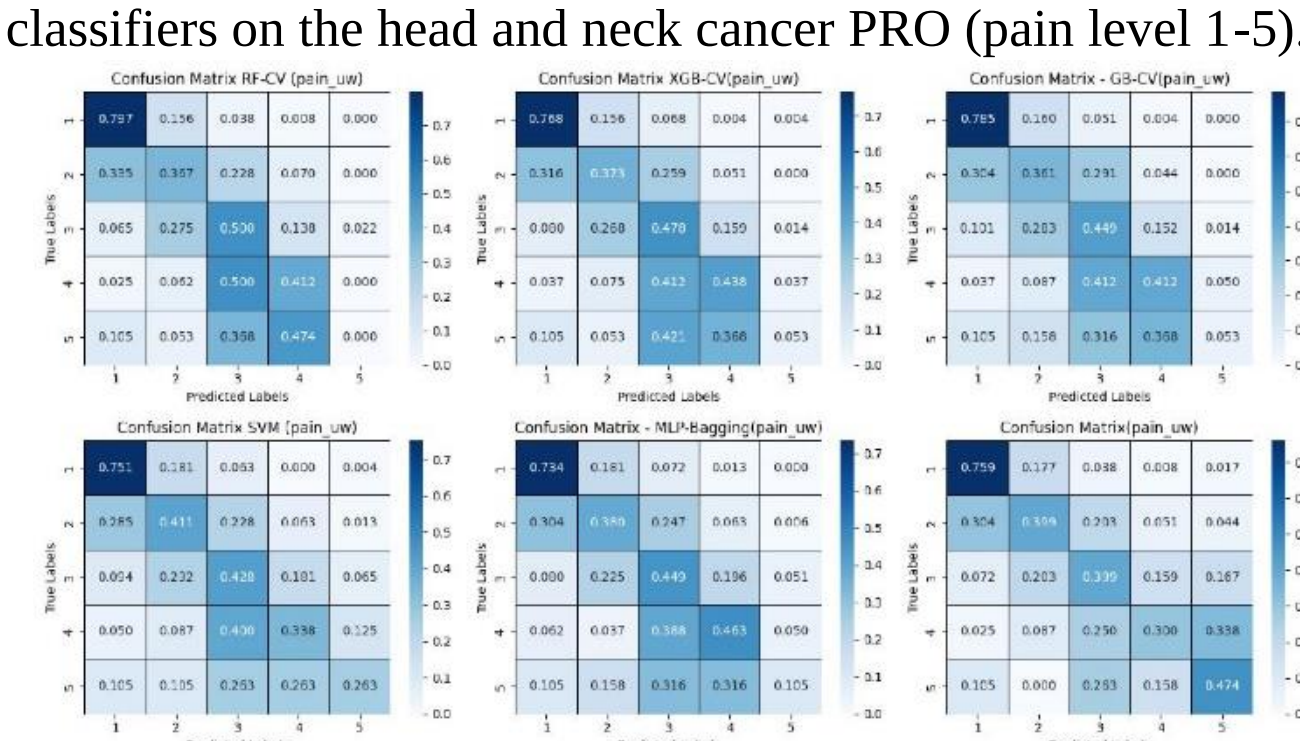

**Fig. 3.** Confusion matrices of six different machine learning classifiers on the head and neck cancer PRO (pain level 1-5).

### *E. ROC Comparison*

Figs. 2, 5, and 8 display the ROC curves for each subclass within three distinct cancer datasets. Our analysis revealed that most methods consistently achieve relatively high AUC scores for pain levels 1 and 4. Notably, Random Forest (RF) secures high AUC scores for pain levels 2 and 3, while Logistic Regression (LR) attains the highest AUC scores for pain level 5 when compared to the other five baseline methods. Similar patterns are observed in prostate and breast cancer datasets for pain and sleep discomfort outcomes, respectively. These findings suggest that RF is particularly effective for classifying the majority classes, whereas LR excels in minority classification.

Additionally, the results depicted in Figs. 2, 3, 11, and 12 include comparisons of the ROC curves and confusion matrices both with and without oversampling. These comparisons underscore the effectiveness of deep learning-based methods, such as MLP-Bagging, in managing multi-class classification tasks with skewed data distributions. After implementing oversampling to balance the training set, some traditional methods like LR, SVM, and GB demonstrated enhanced performance relative to the deep learning-based approach, as evidenced by the results presented in these figures.

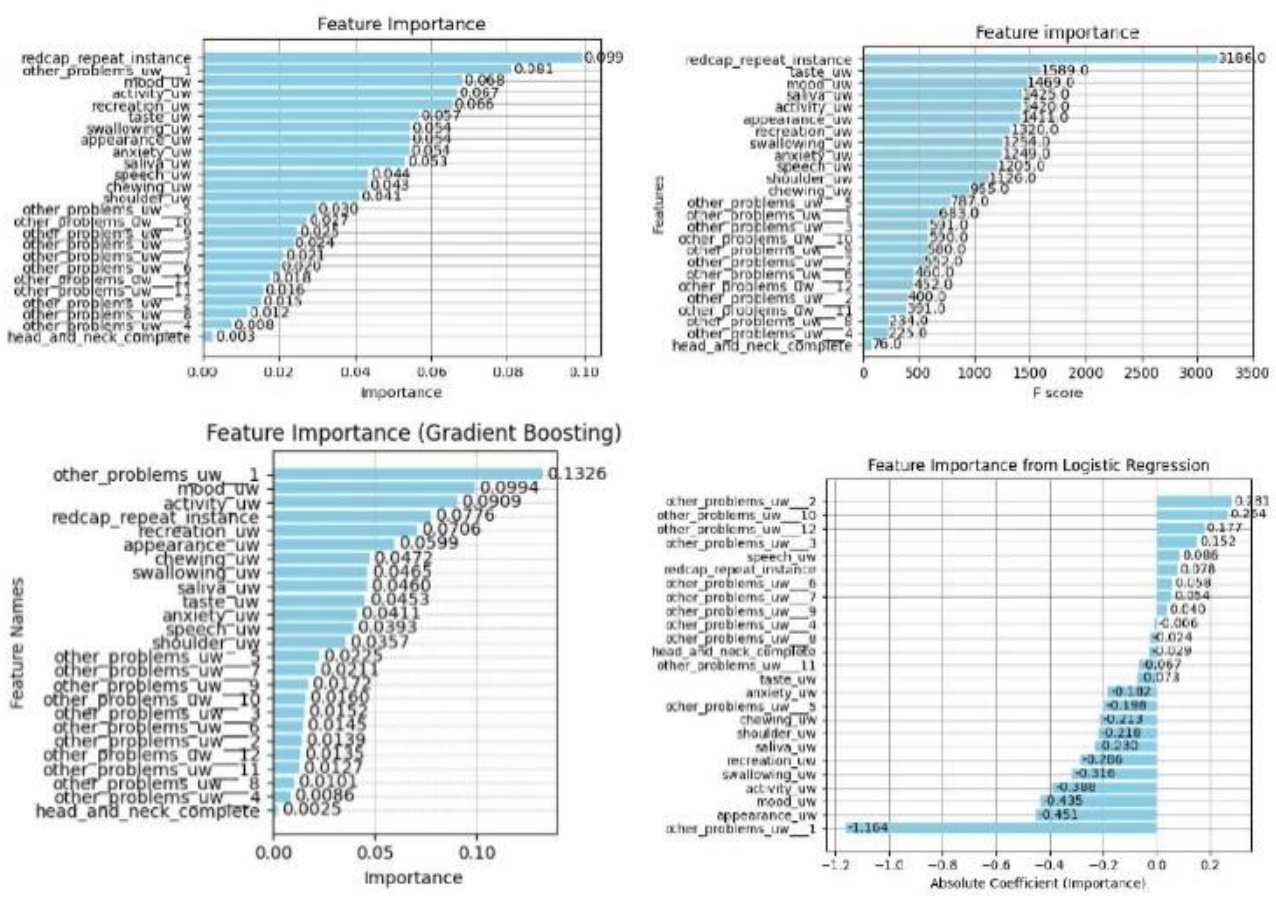

**Fig. 4.** Feature importance scores of four different machine learning classifiers on the head and neck cancer PRO (Top left RF; Top right XGB; Bottom left GB; and Bottom right LR).

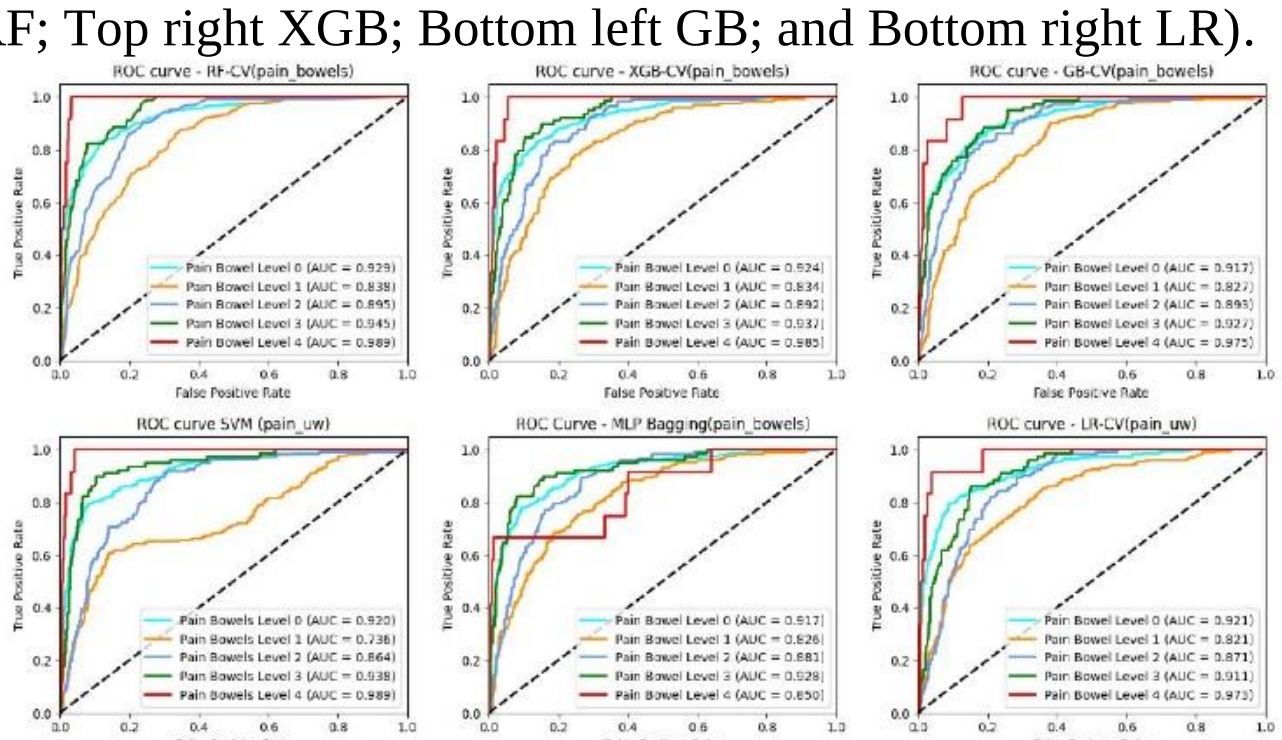

**Fig. 5.** AUC curves of six different machine learning classifiers on the prostate cancer PRO (bowels pain level 0-4).

*F. Confusion Matrix Comparison*

Figs. 3, 6, and 9 display the confusion matrices for each subclass across three cancer datasets. Our analysis shows that nearly all methods attain relatively high classification accuracy for pain levels 1 and 4. Notably, RF demonstrates high accuracy for pain levels 2 and 3, while LR achieves the highest classification accuracy for pain level 5 compared to the other five baseline methods. For the patients with similar pain levels, almost all classifiers suffer a low classification accuracy while a high over or under rate classification (for example pain levels 3 and 4) as demonstrated in the matrix with adjacent positions. This indicates that incorrect evaluation of similar pain levels from patients may increase the difficulty for the classifiers to make the correct prediction. Similar trends are noted in the prostate and breast cancer datasets concerning pain and sleep discomfort outcomes, respectively. These findings reinforce the conclusion that RF is particularly suited for classifying majority classes, whereas LR excels in minority class classification among the baseline methods.

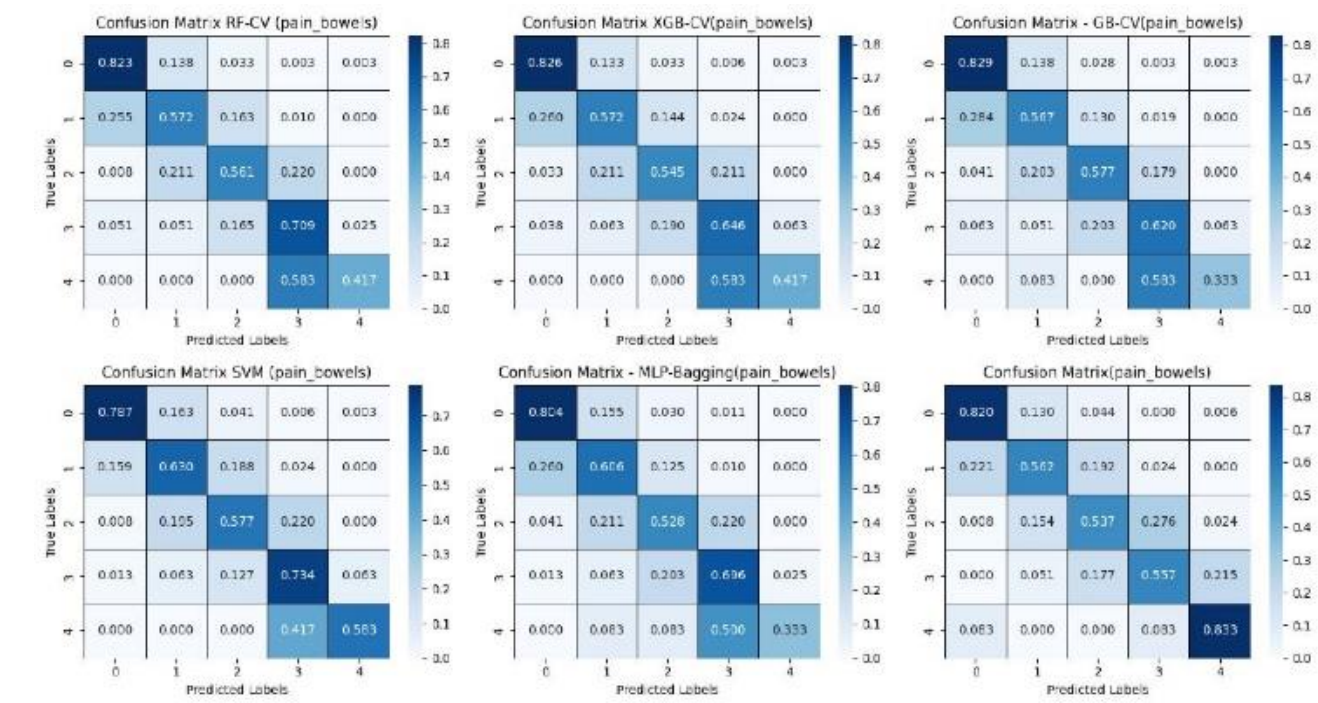

**Fig. 6.** Confusion matrices of six different machine learning classifiers on the prostate cancer PRO (bowels pain level 0-4).

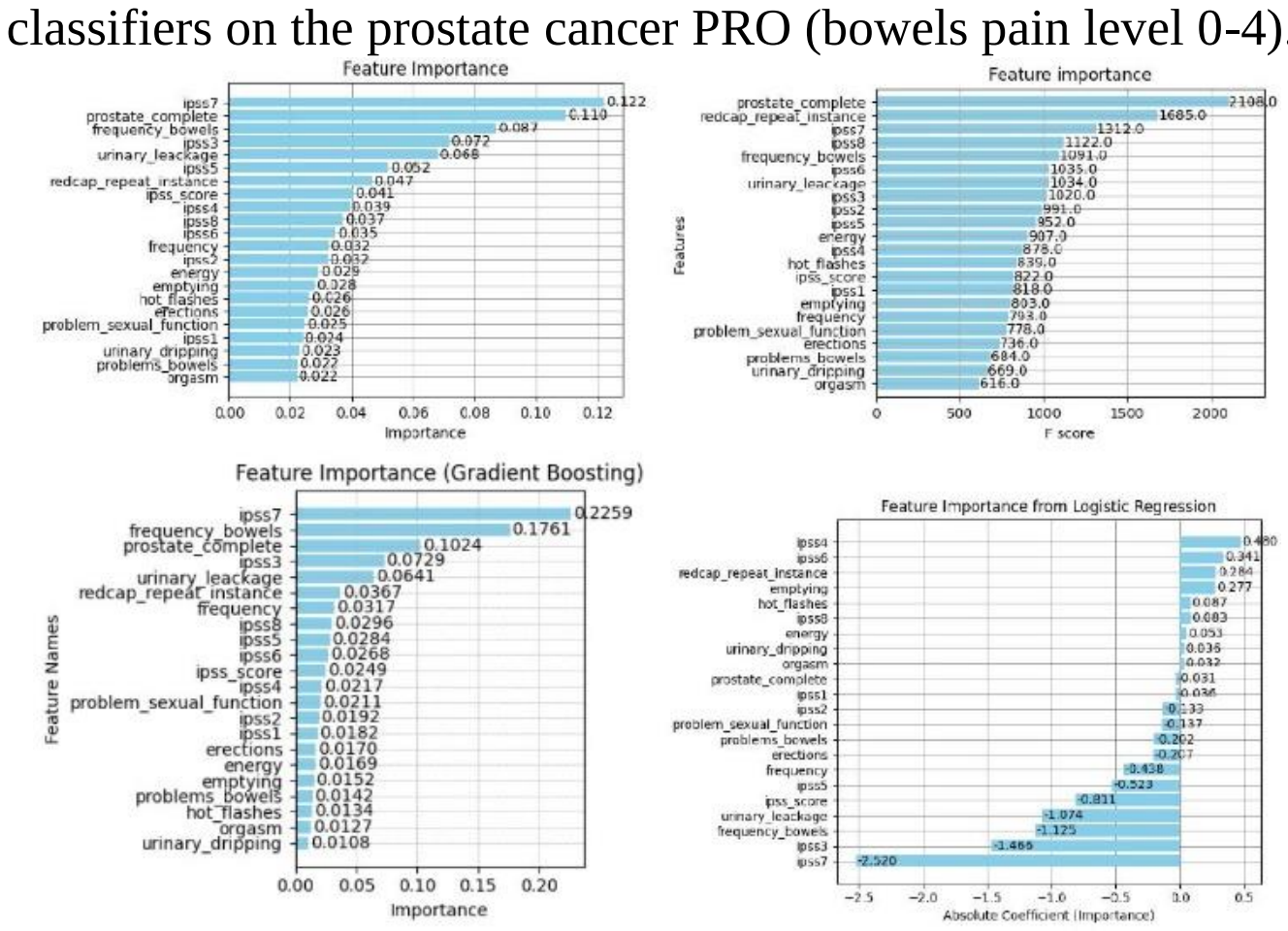

**Fig. 7.** Feature importance scores of four different machine learning classifiers on the prostate cancer PRO (Top left RF; Top right XGB; Bottom left GB; and Bottom right LR).

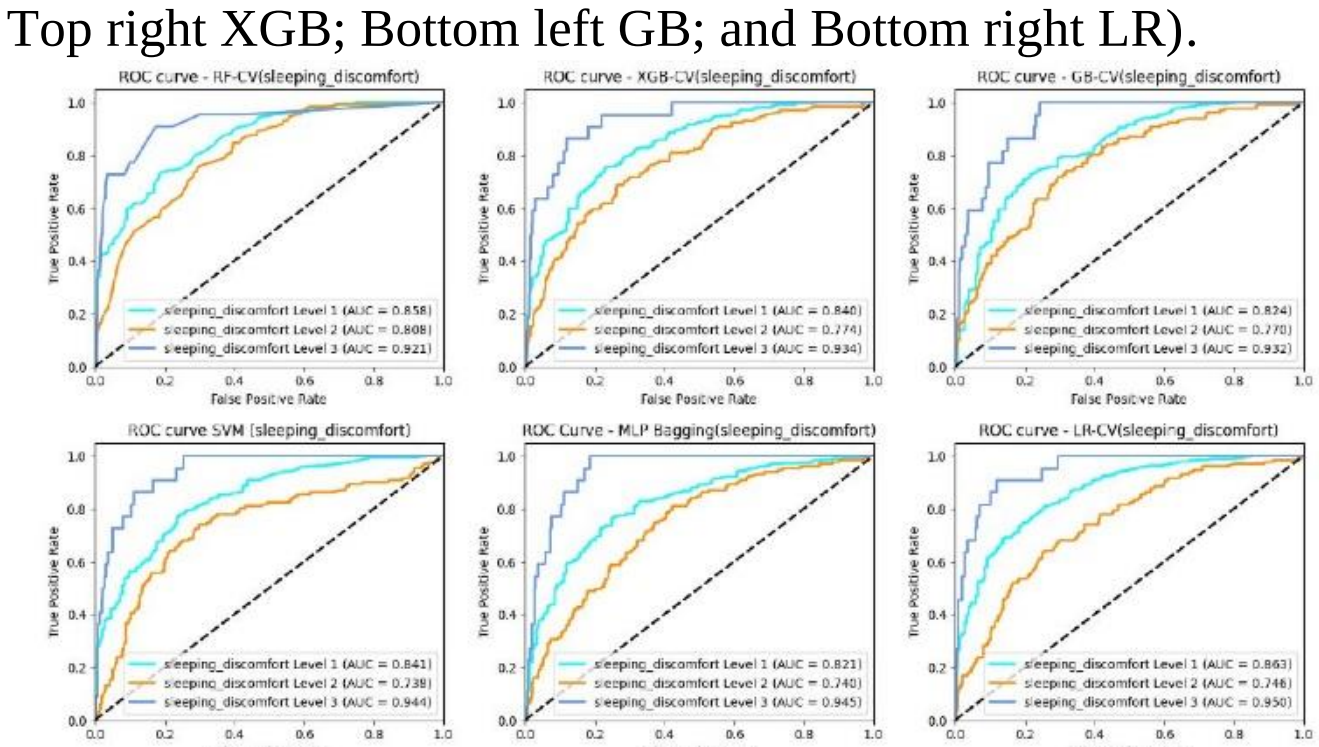

**Fig. 8.** AUC curves of six different machine learning classifiers on the breast cancer PRO (sleeping discomfort 1-3).

Additionally, we examined the importance scores for RF, XGB, GB, and LR. We found that the feature "redcap-repeat-instance" is among the most significant attributes for RF and XGB in enhancing classification performance. The ranking of features along with their respective importance scores or coefficients offers valuable insights into the functioning and utility of these ML classifiers, providing a clearer understanding of their predictive capabilities within the PRO datasets.

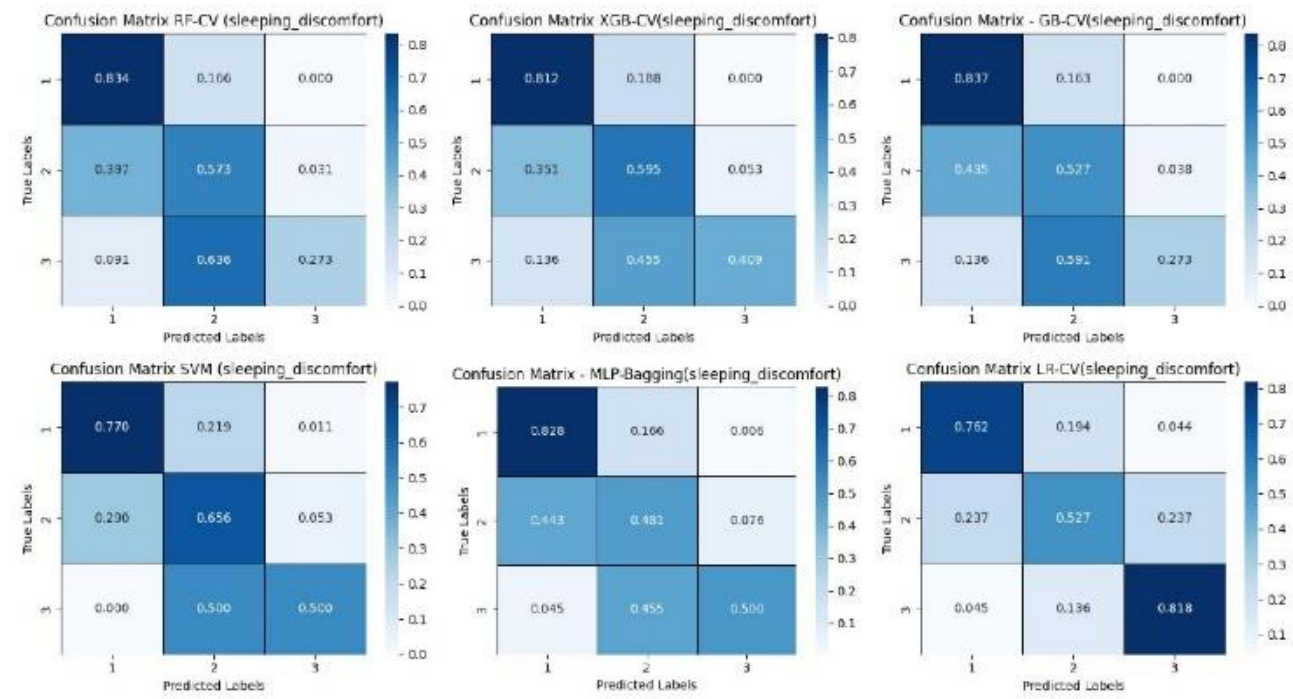

**Fig. 9.** Confusion matrices of six different machine learning classifiers on the breast cancer PRO (sleeping discomfort 1-3).

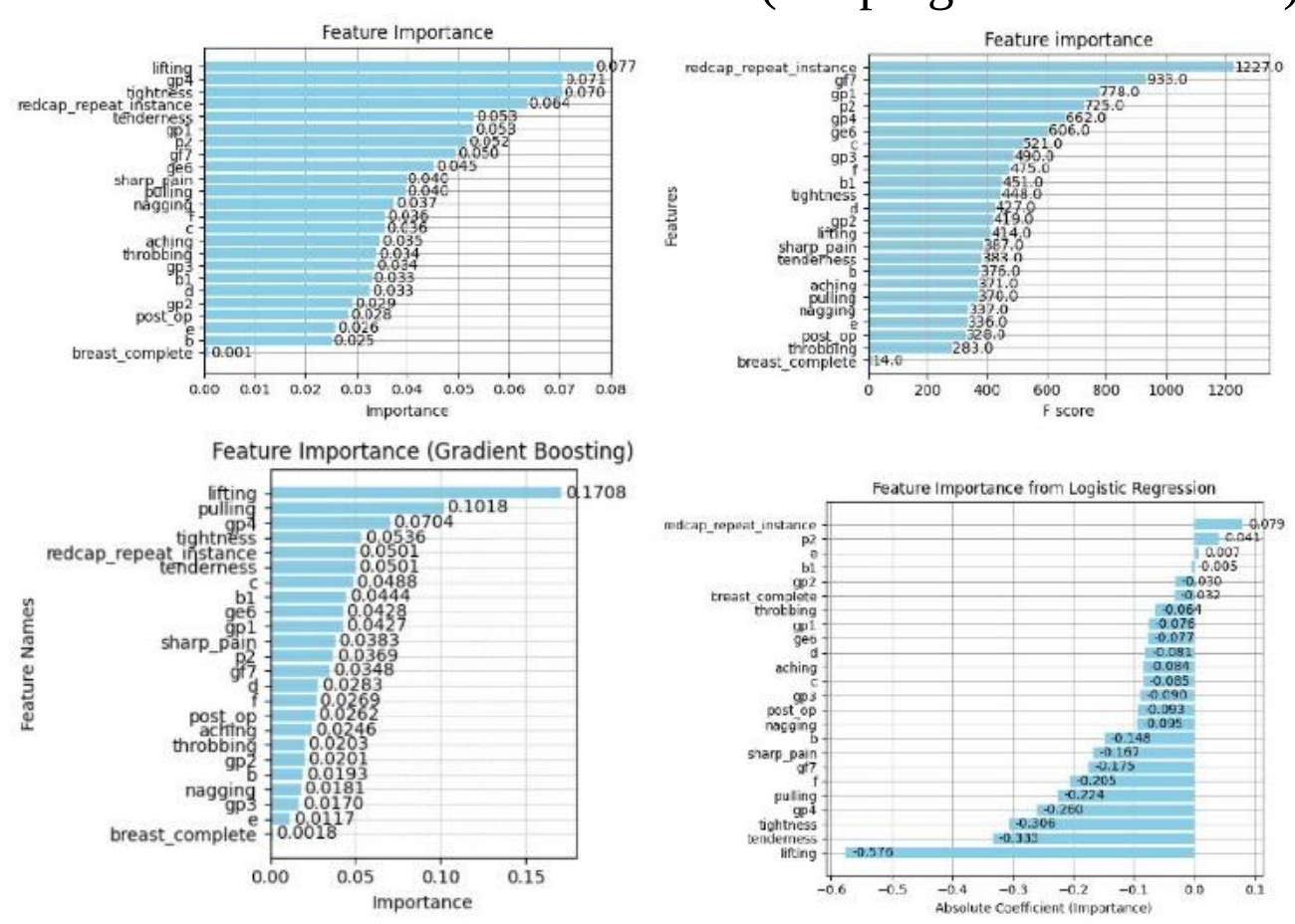

**Fig. 10.** Feature importance scores of four different machine learning classifiers on the breast cancer PRO (Top left RF; Top right XGB; Bottom left GB; and Bottom right LR).

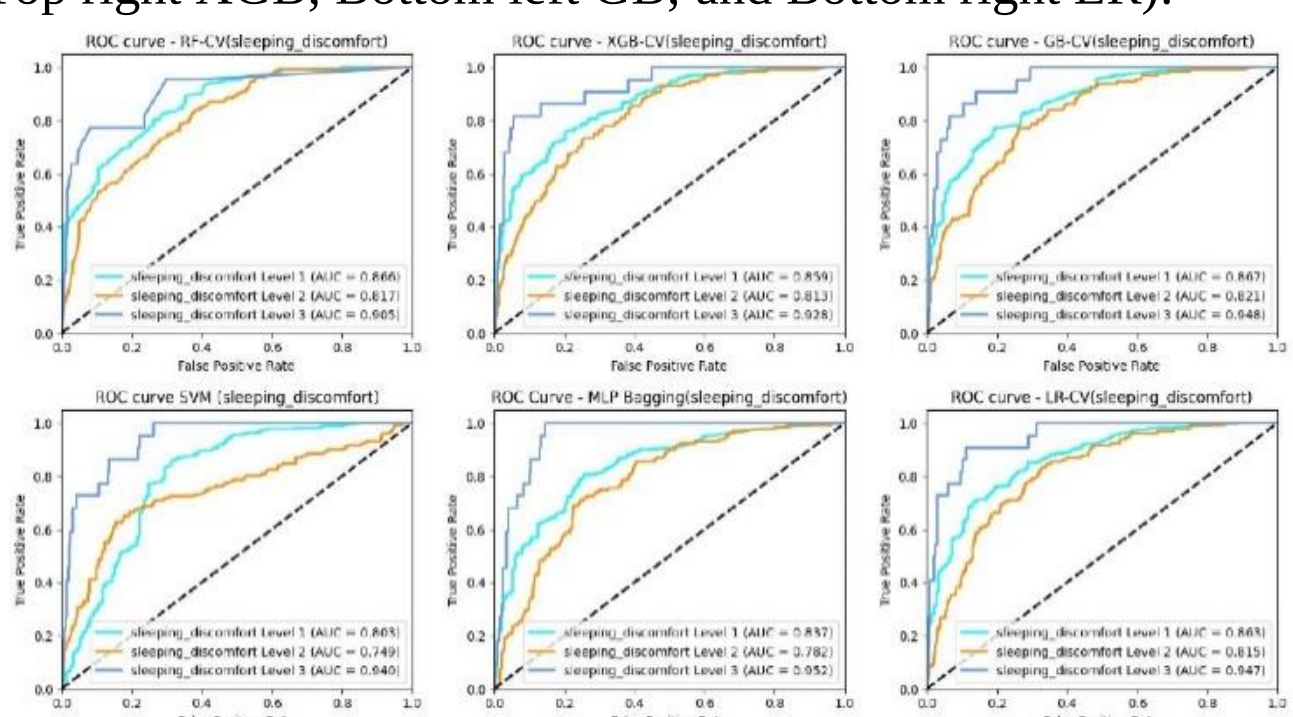

**Fig. 11.** AUC curves of six different machine learning classifiers without oversampling on the breast cancer PRO (sleeping discomfort 1-3).

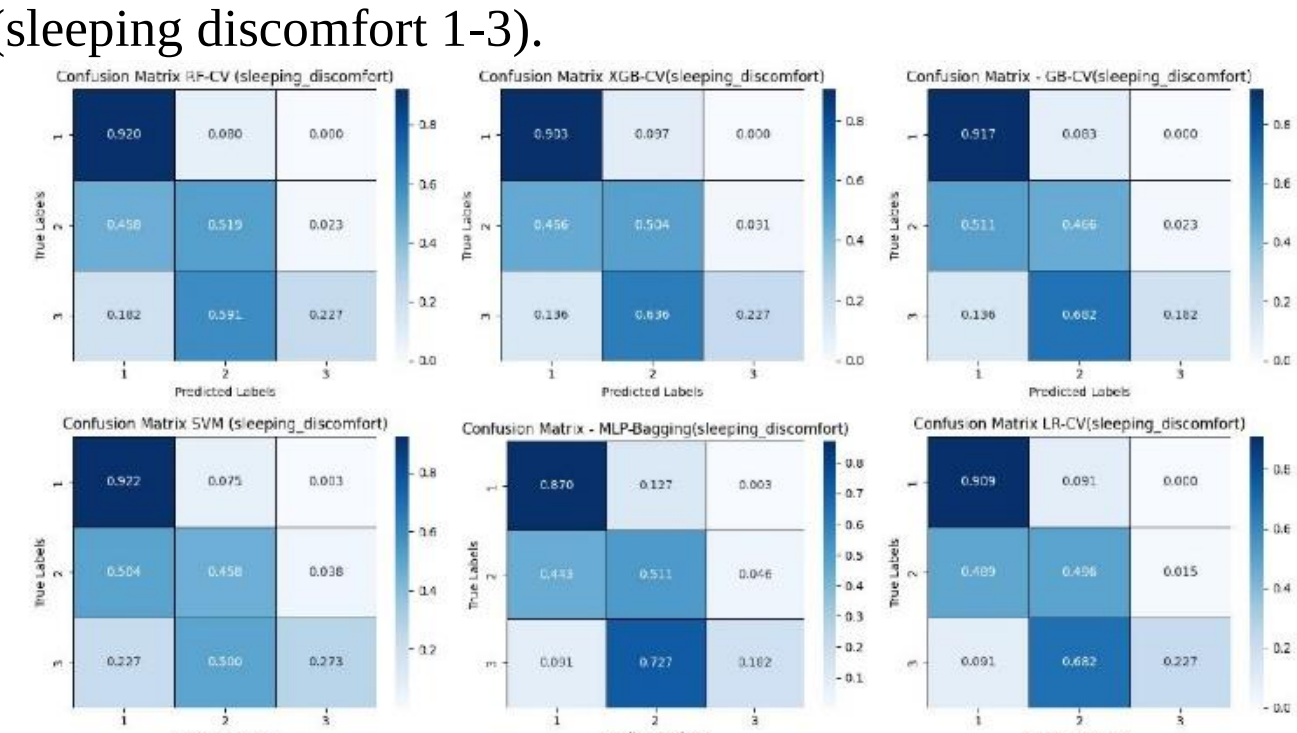

**Fig. 12.** Confusion matrices of six different machine learning classifiers without oversampling on the breast cancer PRO (sleeping discomfort 1-3).

## V. Conclusion

In summary, this study highlights the effectiveness of utilizing multiple ML techniques and strategically addressing imbalanced datasets through oversampling to improve the predictive accuracy of clinical outcomes (i.e., health status) for cancer patients undergoing radiation therapy. By tailoring classifiers to accommodate various cancer types and the severity of symptoms, we have investigated a set of robust machine learning models for PRO analysis. The RF and XGB models not only enhance the forecasting of patient-reported outcomes but also bolster clinical decision-making through improved interpretability. LR shows great potential to address patient's severe health status with improved classification performance. Our proposed imputation and oversampling approaches demonstrate a significant improvement than the non-oversampling approaches, potentially establishing a new benchmark for further research in predictive analytics within the field of PRO. As part of future work, we plan to investigate how anomaly detection is performed on the prediction for cancer patients with severe health status undergoing radiation therapy using PRO data. We also plan to investigate semi-supervised anomaly detection approaches to predict severe patient health status with only partially labeled information in the clinical practice.